\documentclass[10pt, a4paper]{article}

\usepackage{lrec-coling2024} 
\usepackage{amssymb}
\usepackage{times}
\usepackage{latexsym}
\usepackage{booktabs}
\usepackage{graphicx}
\usepackage{makecell}
\usepackage{adjustbox}
\usepackage{algorithm}
\usepackage{algorithmic}
\usepackage{blindtext}
\usepackage{enumitem}
\newcommand{\header}[1]{\vspace*{1mm}\noindent\textbf{#1}.}
\title{CO3:  Low-resource  Contrastive Co-training  for \\ Generative   Conversational Query Rewrite\\ \vspace*{.5\baselineskip}}
\usepackage{microtype}
\usepackage{inconsolata}
\usepackage{multirow}

\name{Yifei Yuan\textsuperscript{1}\thanks{Work done when the author was an intern at Alibaba.}, Chen Shi\textsuperscript{2}, Runze Wang\textsuperscript{2}, Liyi Chen\textsuperscript{3}, Renjun Hu\textsuperscript{2}, \\
\large\bf Zengming Zhang\textsuperscript{2}, Feijun Jiang\textsuperscript{2} and Wai Lam\textsuperscript{4}}

\address{\textsuperscript{1} University of Copenhagen, \textsuperscript{2} Alibaba Group, \\ \textsuperscript{3} Nankai Unversity, \textsuperscript{4} The Chinese University of Hong Kong \\
         yiya@di.ku.dk, liyichen@mail.nankai.edu.cn, wlam@se.cuhk.edu.hk\\
         \{deling.sc, yunze.wrz, renjun.hrj, zengming.zhangzm, feijun.jiangfj\}@alibaba-inc.com\\}

\abstract{
Generative query rewrite generates reconstructed query rewrites   using the conversation history while rely heavily on gold rewrite pairs that are expensive to obtain. Recently, few-shot learning is gaining increasing popularity for this task, whereas these methods are sensitive to the inherent noise due to limited data size. Besides, both attempts face performance degradation when there exists language style shift between training and testing cases.
To this end, we study low-resource generative conversational query rewrite that is robust to both noise and language style shift. 
The core idea is to utilize massive unlabeled data to make further improvements via a contrastive co-training paradigm.
Specifically, we co-train two dual models (namely Rewriter and Simplifier) such that each of them provides extra guidance through pseudo-labeling for enhancing the other in an iterative manner. We also leverage contrastive learning with data augmentation, which enables our model pay more attention on the truly valuable information than the noise.
Extensive experiments demonstrate the superiority of our model under both few-shot and zero-shot scenarios.
We also verify the better generalization ability of our model when encountering language style shift.
 \\ \newline  }

\begin{document}

\maketitleabstract

\section{Introduction}
Recent progress in deep learning NLP techniques has witnessed a resurgent interest in developing conversational IR systems~\cite{reddy2019coqa,Choi2018QuACQA}. Among these tasks, conversational query rewrite (CQR) aims to convert an in-context query to a more explicit form given its context history~\cite{Elgohary:Peskov:Boyd-Graber-2019,su2019improving,pan-etal-2019-improving}. The rewritten query is semantically equivalent to the original one but can be understood without referring to the context. The main research challenge in the CQR system is that conversational queries are often very concise. Information omission such as coreference and ellipsis can  often be observed, where concepts in previous turns are easy to be referred back or omitted. Specifically, in the CQR task, for original QA pairs in a conversation, a manually rewritten query is provided. For example, as shown in Table \ref{tab:qe_example}, for the second query $\mathcal{Q}_2$, the term \emph{``her''} in the original query is resolved as \emph{``Beyoncé''} in the rewrite. In the third turn, the omitted  information after the term \emph{``What else''} is completed after rewriting. 



\begin{table}[t]\small

    \caption{\label{tab:qe_example} An example of a CQR system. $\mathcal{Q}$, $\mathcal{Q^*}$, and $\mathcal{A}$ denote the queries, the corresponding rewrites and the answers. Red color denotes the coreference rewrite part and blue denotes the ellipsis rewrite part.}
    \centering
    \begin{adjustbox}{max width=0.48\textwidth}
    \begin{tabular}{cl}
	    \toprule[1.0pt]
	    \midrule[0.4pt]
	    \multicolumn{2}{l}{\textbf{Conversation Contexts}}  \\
        $\mathcal{Q}_1$ & What can you tell me about \emph{\textbf{Beyoncé's}} voice ?\\
        $\mathcal{A}_1$ & Her tone and timbre as particularly distinctive... \\
	   
        $\mathcal{Q}_2$ & What are some other facts about \emph{{\color{red} her}} voice ?  \\
        
        $\mathcal{A}_2$ & The New York Times commented her voice is "velvety yet tart"...  \\
        $\mathcal{Q}_3$ & What else ?  \\
        $\mathcal{A}_3$ & Other critics praises she was "capable of punctuating any beat".  \\
        \midrule[0.4pt]
        \multicolumn{2}{l}{\textbf{Query Rewrites}}  \\
        $\mathcal{Q}_2^*$ & What are some other facts about \emph{{\color{red} Beyoncé's}} voice ?  \\
        $\mathcal{Q}_3^*$ & What else \emph{{\color{blue} can you tell me about Beyoncé's voice }}?  \\
        
	    \midrule[0.4pt]
	    \bottomrule[1.0pt]
    \end{tabular}
    \end{adjustbox}
	\vspace{-0.3cm}
\end{table}

To address the research challenges in the CQR task, generative CQR has gained great research interest recently, which aims to generate high-quality rewrites and formulates it as a standard text generation problem~\cite{Elgohary:Peskov:Boyd-Graber-2019,su2019improving}. However, it has several drawbacks. First of all, traditional generative models often rely on a large amount of gold rewrite data, whose annotation process is often very expensive. In addition, existing few-shot based models are often vulnerable to the inherent noise due to limited data size. Since the quality of well labeled data is vital to the rewrite performance, how to reduce the impact of noise is an important yet underexplored problem. Furthermore, since different annotator writing styles may result in shifted data distribution, a performance degradation occurs when testing cases come from a different data source dissimilar to the training set~\cite{hao2020rast}. 

In this work, we study the generative CQR task under low-resource scenarios. Since pre-trained language models have shown great few-shot and zero-shot learning abilities in many NLP tasks, we develop our model based on pre-trained GPT-2~\cite{Radford2019LanguageMA}. In order to better leverage the large amount of unlabeled data, we propose a co-training paradigm based on iterative pseudo-labeling. Specifically, we aim to train two separate models namely \emph{Simplifier} and \emph{Rewriter} together, where the Simplifier takes the rewritten query as input and outputs the original query and the Rewriter works on the other way round. Both warmed-up by a small amount of labeled data, in each iteration, the two models first make predictions on the unlabeled data, then the pseudo data generated by the Simplifier is used to train the Rewriter and vice versa. Our model leverages the dual nature of the two models and performs iterative training with only unlabeled data, which largely alleviates the heavy cost of obtaining the gold labeled data. Furthermore, by sampling outputs from one model and inputs generated from the other, the paradigm reduces the gap of the distribution between target and output results, thus equipping the model with enhanced robustness when tackling the noise shift in heterogeneous training and testing data. 

To reduce the impact of noise in the input queries, we further enhance the model by employing a contrastive learning based data augmentation strategy. Inspired by~\citet{gao2021simcse}, we augment the input text by passing it to the model twice to obtain two different embeddings with the same dropout rate. We divide the contrastive loss into internal and external parts. The former considers the two augmented embeddings as positive pairs, while the latter takes the average of the two embeddings and the target embedding as positive pairs. This strategy involves more changes to the original data and helps to learn the common semantic features between the similar inputs and distinguish the differences between dissimilar ones. 

We conduct extensive experiments on two datasets. Our model outperforms state-of-the-art methods under both few-shot and zero-shot settings. Furthermore, we investigate the effect of weakly labeled data size on the performance by adjusting the confidence thresholds and enlarging the unlabeled dataset. The results show that the performance can still be improved when the unlabeled dataset is large enough. In addition, to show that our model has better generalization ability than existing methods, we further perform cross training and testing among two datasets.\footnote{The code is available in \url{https://github.com/yfyuan01/CO3}.} 

In conclusion, the main contributions are: (1) We propose a novel framework for generative CQR tasks in low-resource settings. Our framework combines a Simplifier and a Rewriter through iterative pseudo-labeling, leveraging the contrastive co-training paradigm. (2) We employ an effective contrastive learning based data augmentation strategy to distinguish the truly valuable information from the noise in the input. (3) Extensive experiments and analyses are performed to show the effectiveness and the superior generalization ability of CO3 when encountering language style shift.
\begin{figure*}
    \centering
    \includegraphics[width=.92\linewidth,height=55mm]{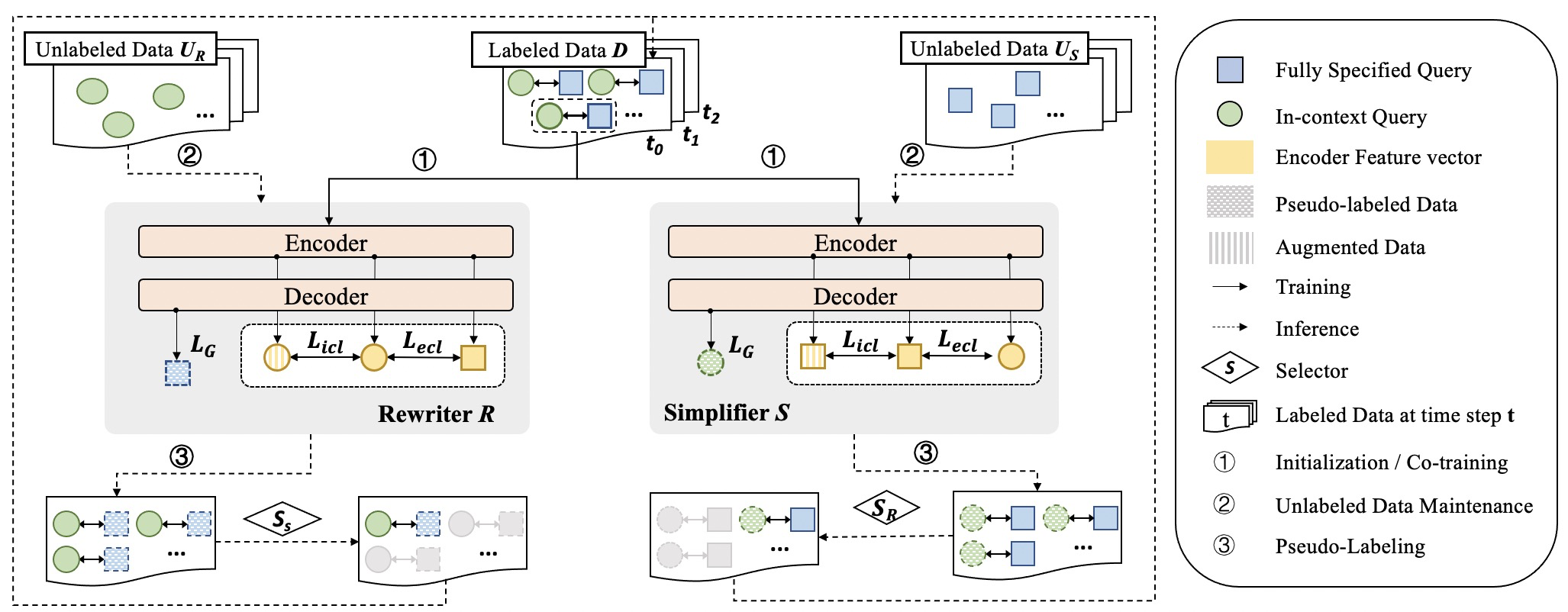}
    \caption{The overall framework of our proposed paradigm. }
    \label{fig:overall}
\end{figure*}

\section{Related Work}
\subsection{Conversational Query Rewrite}
CQR aims to generate explicit rewrites for abbreviated in-context queries~\cite{Tredici2021QuestionRF}. Following this line, many efforts treat this task as a module of the  conversational system, including performing query expansion that selects important terms in the history context~\cite{voskarides2020query,Mele2020TopicPI}, encoding the user’s question in a latent space~\cite{Yu2021FewShotCD}, and contextualizing query embeddings within the conversation~\cite{Krasakis2022ZeroshotQC,Lin2021ContextualizedQE}, etc.

Aiming at generating rewrites that are clear to humans reader, generative CQR treats the task as a standard text generation problem which can be solved via a Seq2Seq model~\cite{Elgohary:Peskov:Boyd-Graber-2019,pan-etal-2019-improving,su2019improving}. Further improvements are made to make the generated rewrite more accurate by developing a multi-task framework~\cite{Rastogi2019ScalingMD,song2020two,zhang2020filling},  incorporating semantic knowledge~\cite{Xu2020SemanticRL,hao2020rast,Liu2020IncompleteUR}, or adding multimodal information~\cite{Yuan2022McQueenAB}. However, these works often rely on large amount of human rewrite data~\cite{Vakulenko2021ACO}, whose  annotation phase is very expensive.
We focus on the generative query rewrite under the low-resource scenario. Under this setting, the work by \citet{yu2020few} is the most relevant,  which proposes two methods named rule based and self-training to transform ad hoc search sessions as pseudo target query rewrites. 

\subsection{Co-training Paradigm}
As an extension of self-training, co-training is a semi-supervised learning technique where two or more models are trained by each other’s predictions~\cite{blum1998combining,abney2002bootstrapping}.  
In NLP areas,~\citet{wan2009co} first proposes a co-training approach to make use of unlabeled Chinese data.~\citet{Wu2018ReinforcedC} focus on the selection of samples and employ a reinforcement learning method to learn a data selection policy with a small labeled dataset.~\citet{Chen2018CotrainingEO} co-train the embeddings of knowledge graphs, whose performance improves at each iteration. In conversation-based tasks, co-training has been employed in the conversation disentanglement task~\cite{Liu2021UnsupervisedCD}. Two neural modules called message-pair  and session classifier are co-trained  with pseudo data built from an unannotated corpus.

\section{Our Framework}
\subsection{Problem Formulation}
Conversational query rewrite aims to reformulate an in-context query to a more explicit form that can be understood without previous context history. Given a conversation context $H$ with $m-1$ turns, it usually consists of several queries and can be denoted as $H=(q_1,q_2,...,q_{m-1})$. Since the queries in the conversation often contain information omission, the task is to generate a rewrite $q^*$ for the query on the latest turn $q_m$ based on $H$. Specifically, in generative CQA, a  query rewriter is trained to generate the de-contextualized rewrites given the conversation history of previous turns
\begin{equation}
\small
    q^* = Rewriter(H,q_m).
\end{equation}

\subsection{Framework Overview}
Figure \ref{fig:overall} depicts the overall structure of our framework. Our co-training framework consists of a \emph{Rewriter} and a \emph{Simplifier} with dual nature. The Rewriter generates the fully specified rewrite based on the original in-context query while the Simplifier works the other way round.

As shown in Figure \ref{fig:overall}, the whole paradigm is contained in a co-training loop where the Simplifier and Rewriter are trained together. The paradigm can be divided into three main steps. The first step \textcircled{1} is an initialization step where the Simplifier and Rewriter are warmed-up by a small number of labeled data $D$. At step \textcircled{2}, we maintain two unlabeled data pools, including the unlabeled Simplifier dataset $U_S$ and the unlabeled Rewriter dataset $U_R$. After that, at step \textcircled{3}, the Simplifier and Rewriter predict and generate weakly labeled data on the unlabeled dataset respectively. The generated weakly labeled data is then fed into a Selector ($S_S$ and $S_R$) which helps to filter the most confident subset of unlabeled samples for better training the models. In the next iteration, the filtered subset is then combined together with existing labeled data to form a synthetic dataset $P$ and is augmented by a contrastive learning based strategy, where all the augmented data is later used to co-train the two models iteratively in the co-training step \textcircled{1}.

In order to enable our model to pay more attention on the truly valuable information, we enhance the model with a contrastive learning based data augmentation strategy, as shown in the yellow part within the Simplifier/Rewriter. Specifically, an in-batch contrastive loss is adopted which contains two parts. The internal part takes the two embeddings of the same sentence  by feeding it to the encoder twice as positive pairs. The external part aims to shorten the pair-wise distance between model outputs and ground-truth rewrites while distinguish the differences between unpaired ones. 

\subsection{Co-training Paradigm}
\header{Model Initialization}
We give a detailed description about the Simplifier and Rewriter in our framework. Both models can be initialized by generative models such as GPT-2~\cite{Radford2019LanguageMA}.

\textbf{Simplifier} is designed to transform the fully specified queries $q^*$ into the simplified original in-context version $q$. Specifically, some terms or specific parts in the input queries may be replaced with pronouns or omitted in the rewrites. For example, after being simplified, the query ``\emph{What empires survived the Bronze Age collapse?}'' is converted into ``\emph{What empires survived?}''. 

\textbf{Rewriter} learns to ``put context back'' to the contextual queries that contain coreference or ellipsis. It is the model we eventually wish to have and is the only model used during the inference stage. 

In the few-shot setting, we use a small amount of well-labeled data to warm up and initialize the two models. Both Simplifier and Rewriter are initialized by the same set of data. For a small labeled dataset $D$, each data sample can be represented as $(H,q,q^*)$, where $H$ is the conversation history, $q$ is the original query, $q^*$ is the gold rewrite. The dataset can be directly used to train the Rewriter. By reversing the source and target query, the Simplifier can be trained inversely from $q^*$ to $q$. In the zero-shot setting, since the gold labeled data is not available, we instead use some weakly labeled data to warm up and initialize the two models. The weakly labeled data is obtained by manually applying some pre-defined rules on the fully specified unlabeled dataset $U_S$. The rules are defined in the same way as \cite{yu2020few}, including replacing some noun phrases with pronouns, etc.

\header{Unlabeled Data Maintenance}
We maintain two additional unlabeled data pools including the unlabeled Simplifier dataset $U_S$ and the unlabeled Rewriter dataset $U_R$. Each data sample in the unlabeled Simplifier dataset $U_S$ is a user search log that contains several fully specified queries. In comparison, the unlabeled Rewriter dataset $U_R$ contains real conversations where each query is contextual. Details of the datasets are described in Section \ref{dataset}.

\header{Pseudo-labeling} After warming up the two models, the Simplifier and Rewriter then make predictions on unlabeled dataset $U_S$ and $U_R$ respectively. For each data input $q_s \in U_S$ and $q_r \in U_R$, the two models generate weakly labeled data as $q'_r$ and $q'_s$. We then use a Selector to filter out predictions with low confidence by setting two confidence thresholds. We set the confidence score of the generated data as the generation likelihood score of both models. The two pseudo-labeled datasets are then fused together to form a synthetic dataset $P$ for further training the model.

\header{Model Co-training} The synthetic dataset $P$ together with the labeled dataset $D$ are used to train a better Simplifier and Rewriter model in the next iteration. Since the well-labeled data is limited and hard to obtain, we hope that the large amount of weakly labeled data helps the model learn the common patterns of the input queries. Before training the model, all the training data is augmented via a contrastive learning strategy which we will describe in detail in the next section. To avoid over-fitting, the two models are reinitialized in every iteration. For the Simplifier, we feed the well specified queries to obtain the abbreviated version and the  Rewriter aims to put context back to provide the rewrites. After training, at the end of each iteration, the models of the next iteration will be overdriven by the newly trained models. Detailed algorithm is shown in Appendix \ref{appendix:1}.

\header{Generation Loss}
The Simplifier and Rewriter both adopt the standard generation loss as the basic training loss. At each time step $j$, the decoder output is determined based on the generated sentence at the previous time steps $y_{<j}$. We minimize the negative log-likelihood of generating the target sentence $y$ given the input $x$ and context history $H$

\begin{equation}
\small
    L_G = {\rm min}\; -\sum^{|y|}_{j=1}logP_\theta(y_j|y_{<j},x,H),
\end{equation}
where $|y|$ is the length of the generated sentence. For Simplifier, the generated $y$ is the simplified original query $q$, while for Rewriter, $y$ is the fully specified query $q^*$.
\subsection{Contrastive Data Augmentation}
We propose a simple but effective contrastive learning based data augmentation method. Motivated by SimCSE~\cite{gao2021simcse}, we also pass the same input to the encoder twice to get two embeddings as positive pairs. Originated from the same sentence, the two embeddings differ in random dropout mask which can be seen as a special data augmentation form. After the dropout augmentation, a contrastive loss is added which takes two embeddings and the ground-truth rewrite embedding as input. Since the dropout can be viewed as a form of noise, this data augmentation technique helps the model learn the shared semantic pattern between the input sentences.

\subsubsection{Internal and External Contrastive Loss}
We divide the overall contrastive loss into internal and external parts. Both of them adopt the same in-batch contrastive loss function (details given in Section \ref{clf}) that takes unpaired samples in a minibatch as negative pairs. 

\header{Internal contrastive loss} 
The internal contrastive loss aims to learn the common semantic features between the similar inputs. It takes the two augmented embeddings originated from the same sentence as positive pairs and aims to equip the model with better capacity to deal with noise. The process can be denoted as
\begin{equation}
\small
    L_{icl} = L_{cl}(Combine[\textbf{Q}';\textbf{Q}'']),
\end{equation}
where $\textbf{Q}'$ and $\textbf{Q}''$ are two query embedding matrices from the same input by feeding into the encoder twice. The $Combine$ function is the concatenation of the two $N \times m$ embedding matrices into one $2N \times m$ matrix with an one-by-one manner. $L_{cl}$ is the in-batch contrastive loss function.

\header{External contrastive loss} The external contrastive loss focuses on shortening the distance between model outputs and the corresponding ground-truth rewrite. Therefore, it takes the average of the two sentence embeddings $Q'$, $Q''$ and the target rewrite $\hat{Q}$ as positive pairs. Similarly, the external contrastive loss can be represented as
\begin{equation}
\small
    L_{ecl} = L_{cl}(Combine[AVG(\textbf{Q}',\textbf{Q}'');\hat{\textbf{Q}}]),
\end{equation}

We denote the final contrastive loss $L_C$ as the sum of   $L_{icl}$ and $L_{ecl}$: $L_C=L_{icl}+L_{ecl}$.
\subsubsection{Contrastive Loss Function}
\label{clf}
For the contrastive loss calculation, we follow the definition given in SimCLR~\cite{chen2020simple}, where the similarity between the representation of an input text and its corresponding positive pair is maximized and the similarity of in-batch unpaired instances is minimized. For a data sample in a minibatch with $N$ instances and its augmented examples with the same size, the corresponding augmented data serves as the positive sample while the remaining $2N-1$ data samples form the negative samples. The contrastive loss function in a minibatch can be represented as
\begin{equation}
\small
    l_{cl}(X_i,X_j)=\frac{exp(sim(X_i,X_j)/\tau)}{\sum_{k=1}^{2N}exp(sim(X_i,X_k)/\tau)},
\end{equation}
\begin{equation}
\small
    L_{cl}(X) =- \frac{1}{2N}\sum_{k=1}^{N}(l_{cl}(X_{2k-1},X_{2k})+l_{cl}(X_{2k},X_{2k-1})),
\end{equation}
where $N$ is the batch size. $X$ is an embedding matrix where the positive pairs in the batch are recorded one by one.
\subsection{Training}
The final loss combines the generation and contrastive loss
\begin{equation}
\small
    L_{all} = L_G + wL_C,
\end{equation}
where $w$ is the contrastive loss weight.

In order to distinguish the well-labeled and weakly labeled data, we also add a weight $\lambda$ for the weak-labeled generation loss. The two types of data are combined by minimizing the loss function
\begin{equation}
\small
    L_G = L_G(D)+\lambda L_G(P),
\end{equation}
\begin{equation}
\small
    L_{all} = L_G(D)+\lambda L_G(P) + wL_C.
\end{equation}

\section{Experiments}
\begin{table}[t]\footnotesize
\setlength{\belowcaptionskip}{5pt}
\caption{The detailed information of the datasets used in our model. w/Omi. denotes if the dataset session contains information omission.}
\label{tab:dataset}
\begin{tabular}{m{1.0cm}cccc}
\toprule
Name    &\textbf{CANARD} &\textbf{TREC} & \textbf{MS MARCO} & \textbf{QUAC} \\ 
\midrule
Session     &    304                                             &            50        &       9306            &         1000      \\ 
Query       &    515                                         &         429            &         13799          &      7354         \\ 
Labeled       &   Yes                                      &        Yes            &         No          &      No         \\ 
w/Omi. & Yes & Yes & No & Yes \\ 
\bottomrule


\end{tabular}
\vspace{-0.4cm}
\end{table}
\label{dataset}
\begin{table*}
\caption{The experimental results on TREC dataset. For GPT-2 based models, we report both results of GPT-2 (base) (within the brackets) and GPT-2 (medium). * denotes that CO3 performs significantly better than other GPT-2 based baselines at 0.05 level using the two-tailed pairwise t-test. $\dag$ denotes the upgraded L-CO3 outperforms all the baselines significantly.}
\label{mainexp}
\begin{adjustbox}{max width=0.99\textwidth}
    \setlength{\tabcolsep}{1.2mm}{
\begin{tabular}{c|c|c|c|c|c|c|c|c}
\hline
   &           Model & \textbf{BLEU-1} & \textbf{BLEU-2} & \textbf{ROUGE-1} & \textbf{ROUGE-2} & \textbf{ROUGE-L} & \textbf{EM}    & \textbf{NDCG@3} \\ \hline
&Original        &     72.50           &    66.17             &        79.71          &    65.66      &      79.66    & 18.65  &    30.40     \\
&Allen Coref &      79.37           &       74.29          &        86.04          &    76.72     &   85.94      & 36.13      &   43.59     \\ 
 \hline
\multirow{5}{*}{\makecell[c]{Zero-\\shot}} &GQR    &        16.02           &        10.63          &       27.37           &      13.13   &   27.29     &  1.47 & 12.56\\
&GPT-2        &     15.41 (15.45)            &        10.54 (10.40)         &       27.17 (28.46)           &      12.42 (12.86)   &   26.75 (28.12)      &  
1.17 (1.86)    &  11.32 (11.56)     \\
&MS MARCO         &     35.19 (34.62)           &    19.90 (19.73)            &       31.06 (29.93)           &   13.18 (13.21)      &   30.41 (29.39)      &    0.93 (0.93)   &  16.90 (14.32)      \\
&Rule Based         &    82.49 (79.31)             &     74.29 (72.30)            &        82.92 (82.93)         &    71.03 (70.53)    &   81.55 (81.86)     &   25.87 (26.81)   & 43.72 (43.25)       \\
&CO3  &     \textbf{83.94*} (\textbf{80.91})            &         \textbf{75.36*} (\textbf{73.37})        &            \textbf{84.08*} (\textbf{83.08})      &     \textbf{72.32*} (\textbf{71.31})    &       \textbf{82.94*} (\textbf{82.02})  &   \textbf{27.91*} (\textbf{27.04})   &    \textbf{45.72*} (\textbf{44.67})     \\
&L-CO3 & 89.42$^\dag$ & 77.31$^\dag$ & 89.06$^\dag$ & 74.90$^\dag$ & 85.26$^\dag$ & 30.55$^\dag$ & 48.90$^\dag$\\
\hline
\multirow{6}{*}{\makecell[c]{Few- \\shot}}&Seq2Seq         &     72.11            &     62.47            &       78.75           &   65.61      &    78.02     &  6.45     &   20.42     \\
&GQR         &        84.84         &        78.80         &      87.42            &     77.93    &    86.40     &   40.82    & 47.28       \\

&GPT-2        & 84.61 (83.20)                &  78.62 (77.00)              &      87.27 (85.52)            &   77.86 (75.79)      &   86.25 (84.66)      &   40.79 (35.89)    &   46.74 (43.28)     \\
&Rule Based         &      85.71 (82.35)          &   79.66 (76.23)              &         88.08 (85.91)        &    78.71 (75.97)    &    86.97 (85.09)    &    40.79 (36.13)  &  49.21 (46.76)      \\
&Self-Learn         &      85.12 (\textbf{83.53})           &    79.73 (77.51)            &         88.22 (86.82)        &   79.36 (76.90)     &    87.38 (85.91)     &  43.12 (38.23)    &   49.24 (46.53)     \\
&CO3 & \textbf{85.87*} (83.42)        &         \textbf{80.24*} (\textbf{78.14})        &          \textbf{89.04*} (\textbf{86.95})        &   \textbf{80.08*} (\textbf{77.48})      &     \textbf{87.92*} (\textbf{86.36})    &     \textbf{44.05*} (\textbf{40.79})  &   \textbf{50.43*} (\textbf{48.26})     \\ 
&L-CO3 & 90.05$^\dag$ & 86.47$^\dag$ & 93.26$^\dag$ & 85.28$^\dag$ & 92.43$^\dag$ & 49.07$^\dag$ & 56.22$^\dag$ \\
\hline
\end{tabular}}
\end{adjustbox}
\end{table*}

\begin{table*}[h!]
\caption{The experimental results of our model compared with the baseline models on CANARD dataset.}
\label{mainexp2}
\begin{adjustbox}{max width=0.99\textwidth}
    \setlength{\tabcolsep}{1.2mm}{
\begin{tabular}{c|c|c|c|c|c|c|c|c}
\hline
        & Model        & \textbf{BLEU-1} & \textbf{BLEU-2} & \textbf{BLEU-4} & \textbf{ROUGE-1} & \textbf{ROUGE-2} & \textbf{ROUGE-L} & \textbf{EM}     \\ \hline
&Original        &     48.86          &    43.04       &  34.43   &        67.98          &    50.58      &      67.91    & 6.99     \\
&Allen Coref &      50.26         &       44.15          &       35.06          &    68.93     &   51.80      & 68.70   & 8.74   \\ 
 \hline
\multirow{5}{*}{\makecell[c]{Zero-\\shot}}&GQR & 9.07 & 5.64 & 2.34 & 15.83 & 4.61 & 14.79 & 0.18\\
&GPT-2        &     10.92 (11.99)            &    5.93 (6.81)         &       2.43 (3.08)           &      15.10 (16.70)   &   4.50 (5.48)      &   13.95 (15.46) &  0.19 (0.39) 
\\

&MS MARCO         &      23.40 (23.42)        &     12.61 (11.95)            &       5.15 (4.44)           &   24.58 (23.29)      &   9.72 (8.87)     &    23.99 (22.70) &   1.94 (1.03)  \\
&Rule Based         &    52.25 (49.15)             &     41.83 (\textbf{40.71})            &        29.53 (29.77)         &    56.43 (58.73)    &   39.40 (41.41)     &   54.57 (57.88) & 3.11 (2.14)       \\
&CO3  &     \textbf{53.21*} (\textbf{49.39})            &         \textbf{42.73*} (40.31)        &            \textbf{30.80*}  (\textbf{30.34})      &     \textbf{59.25*} (\textbf{59.10})    &       \textbf{42.27*} (\textbf{42.73})  &    \textbf{58.40*} (\textbf{57.94})   & \textbf{3.74*} (\textbf{3.55})  \\ 
&L-CO3 & 58.63$^\dag$ & 46.66$^\dag$ & 37.57$^\dag$ & 69.10$^\dag$ & 52.01$^\dag$& 70.33$^\dag$ & 9.47$^\dag$\\
\hline
\multirow{6}{*}{\makecell[c]{Few-\\shot}} 
&Seq2Seq & 45.21 & 37.32 & 26.09 & 53.10 & 38.21 & 54.25 & 3.77\\
&GQR         &        48.03        &     41.20            &      30.98            &    56.72     &  44.10       &      58.82 & 7.84      \\ 
&GPT-2        & 47.52 (47.23)                &  40.01 (38.80)              &      30.34 (28.50)            &   55.59 (54.34)      &   42.38 (38.92)      &   58.76 (54.34)  & 4.66 (3.88)         \\
&Rule Based         &      55.05 (52.07)          &   46.72 (44.48)              &         35.70 (34.54)        &    65.36 (63.41)    &    48.66 (46.99)    &    64.40 (62.42)  & 7.96 (6.80)       \\
&Self-Learn        &      55.77 (52.06)           &    47.40 (44.36)            &         36.15 (34.29)        &  65.84 (63.08)     &    48.86 (46.41)     &  64.75 (61.96)   & 7.57 (7.18)    \\
&CO3  &         \textbf{57.55*} (\textbf{54.83})        &         \textbf{48.55*} (\textbf{46.37})        &          \textbf{36.94*} (\textbf{35.33})        &   \textbf{66.59*} (\textbf{64.85})      &     \textbf{49.35*} (\textbf{47.94})    &     \textbf{65.68*} (\textbf{62.66})  & \textbf{9.02*} (\textbf{8.18})     \\ 
&L-CO3 & 64.29$^\dag$ & 55.46$^\dag$ & 41.73$^\dag$& 72.50$^\dag$ & 55.28$^\dag$ & 74.21$^\dag$ & 12.33$^\dag$\\
\hline
\end{tabular}}
\end{adjustbox}
\end{table*}
\subsection{Datasets}
In our work, both labeled and unlabeled data are used. The information of each dataset are presented in Table \ref{tab:dataset}.
\paragraph{Labeled Dataset.} We perform experiments on two different labeled datasets. First of all, we use the TREC CAst conversational search benchmark~\cite{dalton2020trec}. It contains 50 conversational sessions and 479 queries, each associated with a manual rewrite. In addition, we perform experiments on another query rewrite dataset named CANARD ~\cite{Elgohary:Peskov:Boyd-Graber-2019}. To make it fit to the low-resource setting, we randomly sample 15\% of the original dev set (originated from the QUAC training set) which contains 515 query-rewrite pairs.

\paragraph{Unlabeled Dataset for Simplifier.} For Simplifier, the goal is to generate the simplified version of a fully specified query. We use the ad hoc search sessions collected from MS MARCO~\cite{Campos2016MSMA} directly. Based on the original MS MARCO QA dataset, the artificial search sessions are created using embedding similarity. Each query in the session is consistent with other queries in semantics without any  information omission. We then filter the question-like search sessions from the original dev set and treat each session as a conversation. The total number of session is 9306  with 13799 different queries. One example session of the dataset is: \emph{What is the australian  flag? || What is the population of australia?}. 

\paragraph{Unlabeled Dataset for Rewriter.} For Rewriter, the queries must be context-aware and contain coreference and ellipsis. We use the Question Answering in Context (QUAC) dev dataset~\cite{choi2018quac} which perfectly fits to our setting. The final unlabeled dataset consists of 1000 unique sessions with 7354 queries in total. Each session and query  have 440 and 6.5 tokens on average respectively. One sample session of this dataset is: \emph{What is the australian  flag? || What is the population of this country?}

\subsection{Compared Methods} 
We compare our model with the following methods:
\label{bld}
\begin{itemize}[leftmargin=*]
\item \textbf{Original}. The rewrite is set to be the same as the input query.
\item \textbf{Allen Coref}~\cite{Gardner2018AllenNLPAD} is used for solving the coreference resolution problem in the query. We use it to generate query rewrites.
\item \textbf{MS MARCO} fine tunes GPT-2 on the MS MARCO dataset via a language modeling task.
\item \textbf{Seq2Seq}~\cite{Elgohary:Peskov:Boyd-Graber-2019} is a neural Seq2Seq model where the encoder-decoder structure is based on bidirectional LSTM~\cite{Bahdanau2015NeuralMT,See2017GetTT}.
\item \textbf{GPT-2}~\cite{Radford2019LanguageMA} is  adopted in both settings. In the few-shot setting, we fine-tune the model via cross-validation. In the zero-shot setting, we generate queries without any fine-tuning.
\item \textbf{GQR}~\cite{Tredici2021QuestionRF} is a generative QR method based on T5-large~\cite{Raffel2020ExploringTL}.
\item \textbf{Rule-Based}~\cite{yu2020few} generates weakly labeled data by setting two simple rules, which create abbreviated query given its full  version.
\item \textbf{Self-Learn}~\cite{yu2020few} provides a method for generating the weakly labeled data. A Simplifier is trained separately and applied to the MS MARCO artificial sessions to generate weakly labeled data.
\item \textbf{L-CO3}~\cite{Touvron2023LLaMAOA} is an upgraded version of CO3, with the base model Llama, to test our model with the support of LLMs. 
\end{itemize}

\subsection{Experimental Settings}
\label{es}
The code of our model is based on PyTorch and Huggingface Transformers~\cite{Wolf2019HuggingFacesTS}. In the few-shot setting, we fine-tune the model with 5-fold cross validation following ~\cite{yu2020few}. We split the sessions of two labeled datasets into five folds, where four are used for training and one is used for testing. With different training and testing portions, the whole process includes five rounds. We report the average score of them. Under the zero shot scenario, the whole dataset is used for the testing without splitting. The training is also conducted for 5 rounds with different random seeds. By default, we set the batch size as 4 and the learning rate as 5e-5. 
The evaluation metrics can be divided according to two aspects. We employ some generation evaluation metrics including BLEU ~\cite{papineni2002bleu}, ROUGE~\cite{lin2004rouge}, and Exact Match (EM) to measure the rewrite quality. In addition, for the TREC CAST dataset, we also report the mean NDCG@3 to evaluate the ranking results with the rewritten query. In detail, the rewritten query is used to retrieve relevant passages with Anserini BM25~\cite{Robertson2009ThePR} toolkit and a BERT~\cite{Devlin2019BERTPO} re-ranker is used to re-rank the candidates.

\subsection{Main Experiment Results}
Table \ref{mainexp} and Table  \ref{mainexp2}\footnote{Some numbers may be slightly different from the original paper because some evaluation codes are not publicly available. We use their model code and our own evaluation metric code to do the testing.} show the main experiment results, we have the following observations:
first of all, pretrained language models have a great few-shot learning ability. Even with small amount of data, the Self-Learn model outperforms the Allen Coref model on the TREC dataset. In addition, in the few-shot setting, fine-tuning GPT-2 with small amount of labeled data has improved the BLEU-2 performance from 62.47 to 78.62 compared with traditional Seq2Seq model in TREC. Besides, under the zero-shot setting, directly using MS MARCO sessions to fine-tune GPT-2 model is far under satisfaction. However, by manually defining some rules, the Rule Based model achieves 82.49 and 52.25 BLEU-1 result in TREC and CANARD, which verifies the importance of the weakly labeled data. CO3 achieves the best overall performance among all the methods on two datasets in both settings. Specifically, the GPT-2 medium based version performs significantly better than other GPT-based baselines. In the few-shot setting, CO3 outperforms the Self-Learn method using the same amount of weakly labeled data. In the zero-shot setting, the superior results also  prove the benefit of our co-training paradigm. Besides, with the help of LLM, the upgraded L-CO3 gains further performance lift. This demonstrates the superiority of our paradigm in the generative LLM era. 
\begin{table}[t]
\caption{Ablation study of our model on TREC dataset, where CL represents contrastive learning.}
\label{abl}
\begin{adjustbox}{max width=0.5\textwidth}
    
\begin{tabular}{c|c|c|c|c}
\hline
 &Model               & \textbf{BLEU-2} & \textbf{ROUGE-L} & \textbf{EM} \\ \hline
\multirow{4}{*}{\makecell[c]{Few-\\shot}} &1.) w/o External CL &          79.96      & 87.40              &    43.25        \\
&2.) w/o Internal CL &             80.03    &       87.44           &  43.69      \\
&3.) w/o CL   &  79.72     &     87.34            &  43.19                          \\
&4.) w/o Simplifier  &      78.90           &          87.43        & 42.90            \\
&CO3 & 80.24 & 87.92 & 44.05 \\ \hline
\multirow{4}{*}{\makecell[c]{Zero-\\shot}}&1.) w/o External CL &    75.02             &    82.63              &    27.56         \\
&2.) w/o Internal CL &      74.88           &            82.31      &      26.80       \\
&3.) w/o CL          &  74.30               &       81.61           & 25.90            \\
&4.) w/o Simplifier  &       74.52          &  81.78                &    26.20         \\
&CO3 & 75.36 & 82.94 & 27.91 \\\hline
\end{tabular}
\end{adjustbox}
\vspace{-0.4cm}
\end{table}

\subsection{Ablation Study}
To evaluate the effect of different components of our framework, we report the performance of our model with several variants. According to  Table \ref{abl}, without the CL method, the performance decreases in both settings. Besides, contrastive loss is more effective in the zero-shot scenario than in the few-shot setting, which proves that using the contrastive learning based data augmentation technique helps to enhance the model when the gold data is not available. Among the two CL losses, the external loss plays a more important role, where the few-shot EM performance drops 0.8 percent without it. Furthermore, without the Simplifier, the few-shot and zero-shot EM performance decreases 1.15 and 1.71  percent, demonstrating the superiority of our co-training paradigm.
\section{Extensive Analysis}
To further evaluate the model capacity, we conduct several analysis to show the great potential and superior generalization ability of our model. 
\subsection{Weakly Labeled Data Scale Analysis} 
We first analyze the impact of the weakly labeled data involved in model training on the performance.
\begin{table}[t]
    \centering
    \caption{Confidence threshold analysis of CO3.}
    \begin{adjustbox}{max width=0.45\textwidth}
    \begin{tabular}{c|c|c|c|c|c}\hline
         &  $s_s$& $s_r$ & \textbf{BLEU-2} & \textbf{ROUGE-L} & \textbf{EM}\\ \hline
    \multirow{5}{*}{\makecell[c]{Few-\\shot}} 
    & 0 &0 &79.16 &87.22 &42.91 \\
    & 50 & 70 & 77.69 &86.68 &41.78 \\
    & 70 & 90& 78.43 & 86.49 & 39.77 \\
    & \textbf{90} & \textbf{110}& \textbf{80.24}&\textbf{87.92}&\textbf{44.05} \\
    & 110 & 130& 78.21 & 87.52 & 44.02\\ \hline
    \multirow{5}{*}{\makecell[c]{Zero-\\shot}}
    & 0 &0 &74.07 &81.70 &20.28 \\
    & 50 & 70 &74.23 &81.84 &22.61 \\
    & \textbf{70} & \textbf{90}& \textbf{75.36} & \textbf{82.94} &\textbf{27.91} \\
    & 90 & 110&71.66 &81.20&25.87 \\
    & 110 & 130& 74.16&80.83&23.08 \\ \hline
    \end{tabular}
    \end{adjustbox}
    \label{tab:ca}
\end{table}
\begin{table}[]
\setlength{\belowcaptionskip}{2pt}
    \centering
    \caption{Dataset scale analysis of our model. Scale denotes the size of $U_S$ and $U_R$ datasets.}
    \begin{adjustbox}{max width=0.46\textwidth}
    \begin{tabular}{c|c|c|c|c|c}\hline
         & Scale &BLEU-1 & BLEU-2 & ROUGE-L & EM\\ \hline
    \multirow{4}{*}{\makecell[c]{Few-\\shot}} 
    & 10k  & 85.64 & 79.26 & 87.81 & 44.11 \\
    & 20k  & 85.91 & 79.98 & 87.99 & 44.20 \\
    & 30k & 86.17 & 80.34 & 88.05 & 44.35 \\
    & 40k &\textbf{86.35} &\textbf{81.02}&\textbf{88.54}&\textbf{44.76} \\ \hline
    \multirow{4}{*}{\makecell[c]{Zero-\\shot}}
    & 10k &83.83 &75.47 &81.98 &26.11 \\
    & 20k & 84.20 &76.23&81.71 &26.47 \\
    & 30k& 84.90 & 76.79 & 82.97 &27.98 \\
    & 40k& \textbf{85.93} & \textbf{77.21} & \textbf{83.77} &\textbf{28.94} \\
    \hline
    \end{tabular}
    \end{adjustbox}
    \label{tab:da}
\end{table}
\begin{figure}[]
\setlength{\belowcaptionskip}{2pt}
    \centering
    \includegraphics[width=\linewidth,height=34mm]{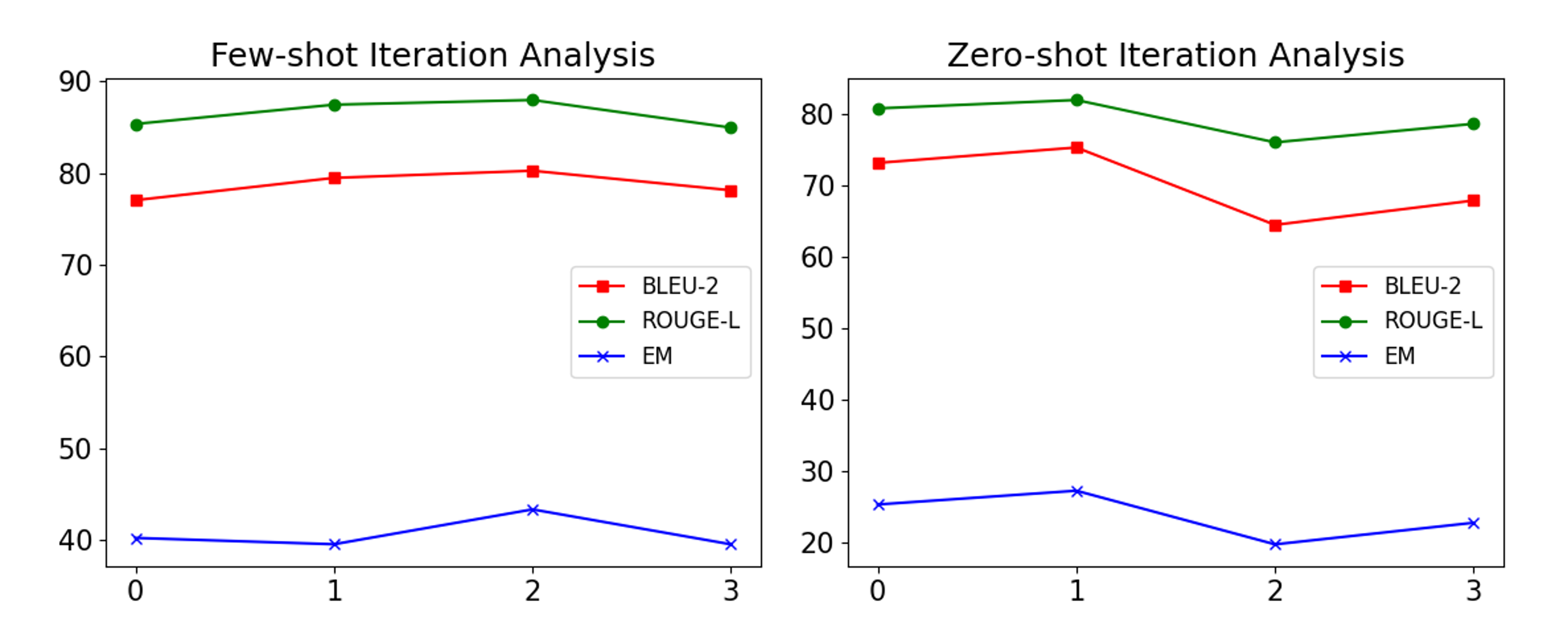}
    \caption{Performance of CO3 in each iteration.}
    \label{fig:ia}
    \vspace{-0.4cm}
\end{figure}
\header{Confidence Threshold Analysis}
We control the amount of training data by adjusting the two numbers and report the results under different Simplifier and Rewriter confidence thresholds.
According to the results shown in Table \ref{tab:ca}, in TREC dataset, the best confidence threshold is lower in zero-shot than in few-shot. The result is because the quality of the generated data is worse in this setting and has a lower score. In addition, starting from zero when we enlarge the two thresholds, the performance first increases while starts to drop at certain stage. This is because when the confidence threshold is too small, large amount of noise data is introduced which negatively affects the rewrite quality. However, when the confidence threshold is too large, most weakly labeled data fails to participate in model training, thus causing model overfitting.

\begin{figure*}
    \centering
    \includegraphics[width=\linewidth,height=60mm]{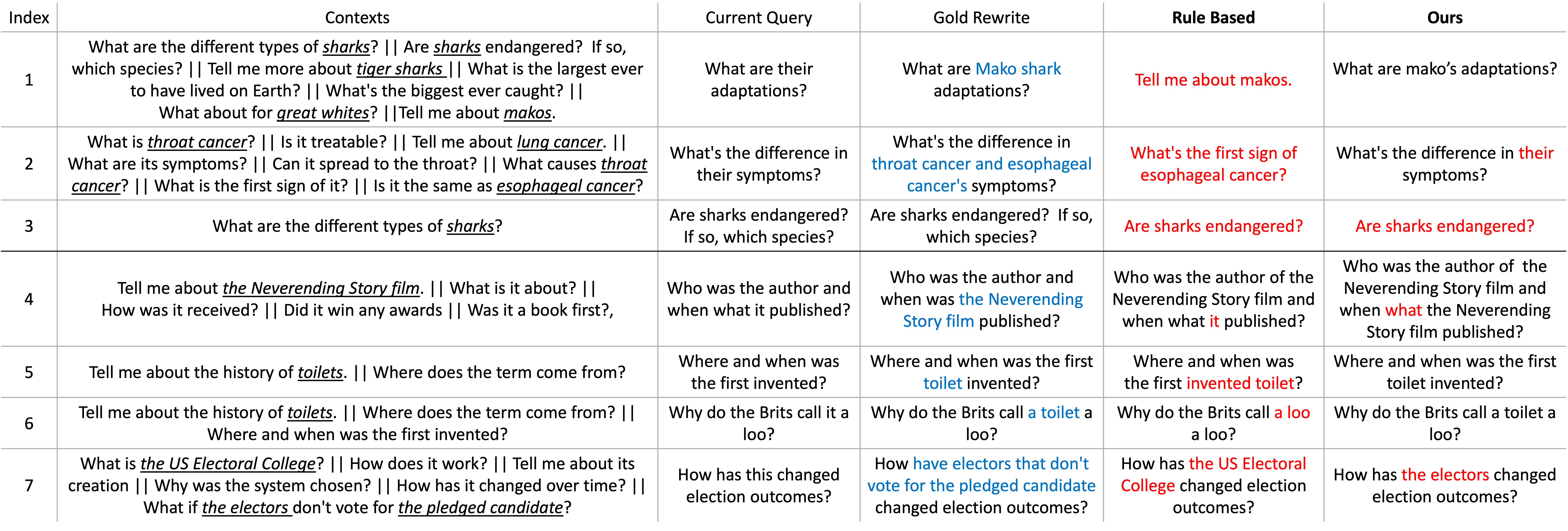}
    \caption{Real case and error case analysis. The first three are examples under the few-shot setting and the last four are under the zero-shot setting. The blue part denotes the resolved coreference or completed ellipsis in the gold rewrite. The red part denotes the errors in the model output.}
    \label{fig:casestudy}
\end{figure*}

\header{Dataset Scale Analysis}
\label{dataset scale analysis}
We fix the confidence  thresholds and incrementally enlarge the unlabeled dataset by adding new instances. 
Table \ref{tab:da} shows the performance under different dataset scales. By enlarging the unlabeled dataset, the performance increases under both settings. Compared with the few-shot setting, dataset scale has more effect on the zero-shot scenario where gold data is unavailable. Notably, when the $U_S$ and $U_R$ dataset reach to 30k and 40k samples, the result exceeds the best performance (zero-shot EM increases from 27.91 to 28.94, few-shot EM increases from 44.05 to 44.76). This proves that we can further improve the performance by setting a high threshold with a larger dataset where large amount of high quality data is filtered to join model training. 

\subsection{Iteration Analysis}
In Figure \ref{fig:ia}, under the few-shot setting, the model performance keeps increasing in the first three iteration and reaches to the best in the third iteration. In the zero-shot setting, the model takes less iterations to reach the peak. It demonstrates that without the gold labeled data, the model is more easy to suffer from overfitting. 
However, compared with traditional methods, our model prohibits data overfitting in two aspects. First, compared with methods with fixed dataset, new data can be introduced to the model in each iteration via  additional weakly labeled data. Second, the dropout based data augmentation strategy makes sure that some random noise is added, ensuring that the data distribution is not strictly alike in each iteration.
\begin{table}[]
\caption{Generalization analysis on two datasets. T, C denote the TREC, CANARD dataset. }
\label{interactive}
\begin{adjustbox}{max width=0.46\textwidth}
\begin{tabular}{c|c|c|c|c|c}
\hline
& Model      & BLEU-1 & BLEU-2 & ROUGE-L & EM \\ \hline
\multirow{4}{*}{\textit{\begin{tabular}[c]{@{}l@{}}T->C \end{tabular}}} & Seq2Seq      &     35.92   &    24.12    &     43.97    & 2.03   \\
& GPT-2      &     50.83   &    43.01    &     59.60    & 4.47   \\
& Rule-Based &54.55        &   44.91     &  59.64       & 5.24   \\
& Self-Learn         &    52.76    &  44.44       &   61.68 & 6.41 \\
& CO3       & \textbf{55.28}       &   \textbf{46.38}     &    \textbf{64.16}     &  \textbf{7.05}  \\ \hline
\multirow{4}{*}{\textit{\begin{tabular}[c]{@{}l@{}}C->T\end{tabular}}}  & Seq2Seq &  68.87      &    59.34    &   75.23      & 5.37  \\ 
& GPT-2      &  80.76      &    72.81    &   79.68      &  29.84  \\
& Rule-Based  &    83.96    &  76.78  &84.37 & 36.13     \\
& Self-Learn   &81.13      &   74.07     &   82.96      & 30.77   \\
 & CO3       & \textbf{84.23}       &   \textbf{77.15}     &   \textbf{85.52}      &  \textbf{38.63}  \\ \hline
\end{tabular}
\end{adjustbox}
\vspace{-0.4cm}
\end{table}
\subsection{Generalization Analysis}
We train our models on one dataset while testing on the other to explore the generalization ability of our model. Table~\ref{interactive} shows that the traditional non-pretrained Seq2Seq model encounters severe performance drop when the testing data is different from the training data, where the TREC EM performance drops to 5.37 when training on the CANARD dataset, and the CANARD performance also decreases to 2.03 when training on TREC. This is mainly due to the writing style shift between the heterogeneous training and testing samples. Compared with raw GPT-2, models enhanced with weakly labeled data show better performance. This proves that large amount of weakly labeled data helps the model learn the common feature among queries that need to be rewritten. 
Our model achieves the best overall scores on all metrics concerning both two datasets, showing the superiority of CO3 on the cross-dataset  robustness. 

\subsection{Loss Function Analysis}
\label{lf}
\begin{figure}
    \centering
    \includegraphics[width=\linewidth]{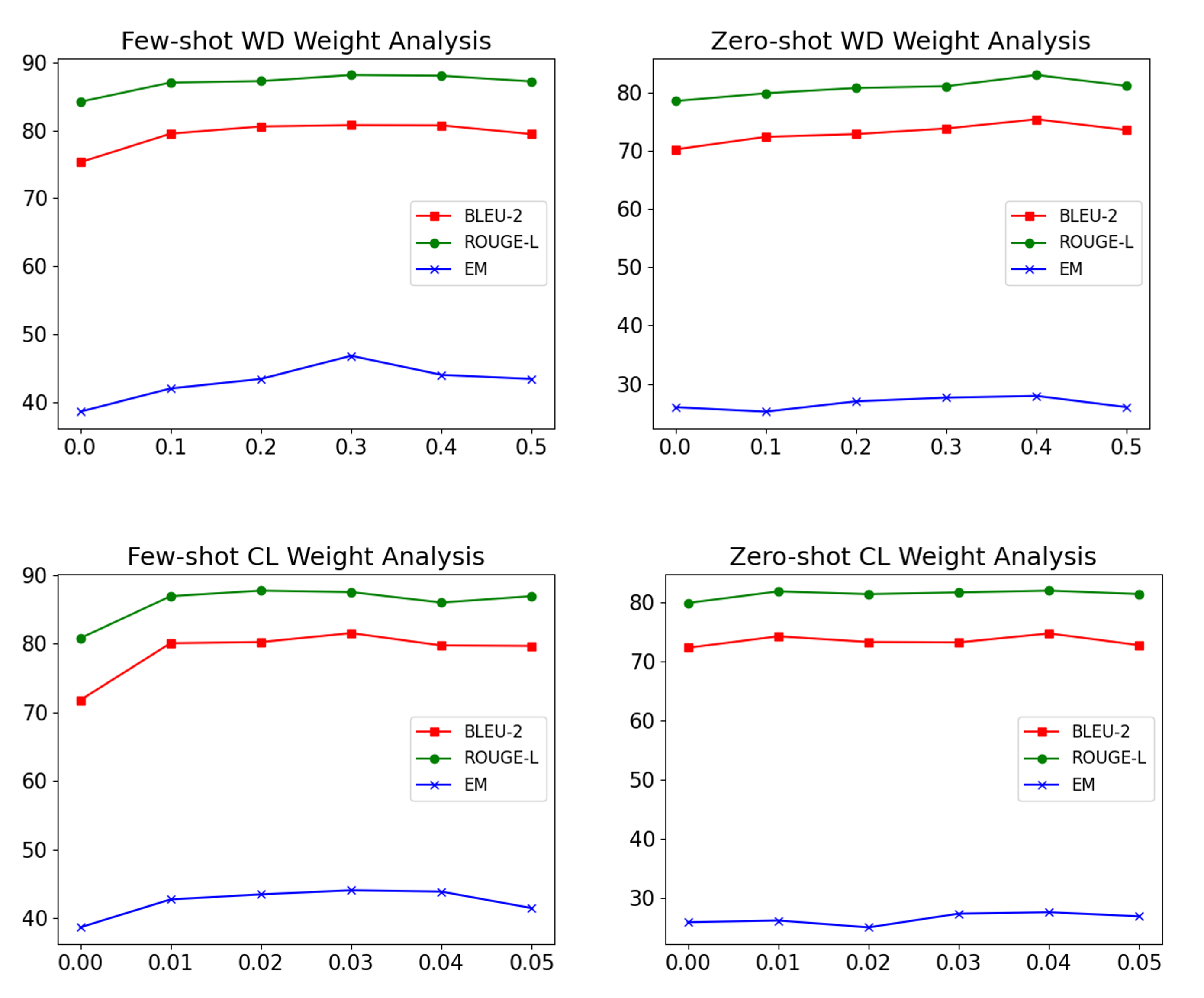}
    \caption{The upper part is the weakly-labeled  data weight analysis and the lower part is the contrastive weight analysis.}
    \label{fig:hyper}
\vspace{-0.6cm}
\end{figure}
\subsubsection{Weakly Labeled Data Weight Analysis}
\label{uwa}
As shown in the upper part of Figure \ref{fig:hyper}, when we increase the weakly labeled data weight $\lambda$, the performance first increases then decreases. The result verifies that although the quality of weakly labeled data may not be as good as the gold rewrite, it has a positive influence on the performance. Furthermore, it can be observed that under the zero-shot setting, the best weakly-labeled data weight is larger than what in the few-shot setting, where the model is more dependant on the large amount of unlabeled data for the lack of gold data for training guidance.

\subsubsection{Contrastive Learning Weight Analysis}
As shown in the lower part of Figure \ref{fig:hyper}, we found that when the contrastive loss number $w$ is around half of the generation loss (around 0.03 and 0.04), the model reaches the best score. We can also have the same observation as Section \ref{uwa}, that is, in the zero-shot setting, the contrastive loss (0.04) under the best performance is slightly larger than that in the few-shot setting (0.03). The result verifies that contrastive loss is more useful when there is few  well-labeled data when training the model and can help the model better tackle the noise.

\subsection{Real Case and Error Case Analysis}
We provide some cases to compare CO3 with the rule-based model and analyze the potential drawbacks. As shown in Figure \ref{fig:casestudy}, CO3 has a better overall rewrite ability. For example, in case 1 and 2, the rule-based method outputs the rewrite of a wrong context sentence instead of the current query that needs to be rewritten. In case 6, the pronoun ``it'' in the original query should refer to ``a toilet'' in the context instead of ``a loo''. However, some errors remain in both models. For instance, in case 3, although the rewrite seems easy and does not require any change, the last sentence is omitted by both models. Besides,  as the conversation goes deeper, coreference that is challenging to both models is more common, such as coreference containing several entities (e.g. case 2) and coreference requiring reasoning between different entities (e.g. case 7).  

\section{Conclusion}
We investigate the conversational query rewrite task under low-resource settings. We propose a co-training paradigm where a Simplifier and Rewriter are jointly trained. The Simplifier takes the fully specified query as input and outputs the abbreviated  query and the Rewriter works the other way round. Based on iterative pseudo-labeling, the two models have dual nature where one takes the output from the other as input in each iteration. To distinguish the truly valuable information of the input, we enhance the model with a contrastive learning based data augmentation strategy. Experiments show the effectiveness of CO3 on two datasets. Extensive analyses are performed to prove the results can be further improved. Future works and limitations are discussed in Appendix \ref{appendix:2}.
\section{Ethics Statement}
In adherence to ethical considerations, our work utilizes exclusively open-source datasets. We have strictly followed all licensing and intellectual property rights associated with these datasets.



\nocite{*}
\section{Bibliographical References}\label{sec:reference}

\bibliographystyle{lrec-coling2024-natbib}
\bibliography{lrec-coling2024-example}

\label{lr:ref}
\bibliographystylelanguageresource{lrec-coling2024-natbib}
\bibliographylanguageresource{languageresource}
\appendix

\section{Co-training Algorithm}
\label{appendix:1}
Algorithm \ref{alg1} shows the detailed algorithm of our co-training paradigm.
\label{sec:alg}
\begin{algorithm}[]
\footnotesize
\caption{Simplifier and Rewriter Co-training Paradigm} 
\label{alg1} 
\begin{algorithmic}[1] 
\REQUIRE~~\\
Simplifier: $S$, Rewriter: $R$\\
Labeled dataset: $D$, Unlabeled Rewriter and Simplifier Dataset: $U_R$, $U_S$\\
Simplifier Confidence Threshold: $s_s$, Rewriter Confidence Threshold: $s_r$ \\

\ENSURE~~\\
A trained Simplifier $S^*$, A trained Rewriter $R^*$\\
\STATE Initialize $S$ and $R$ and train them on $D$ 
\WHILE{$U_S\neq \varnothing$ and $U_R\neq \varnothing$}
\STATE $P_S\leftarrow$[ ] , $P_R\leftarrow$[ ]
\FOR{$q_s\in U_S$}
\STATE $q'_r\leftarrow$ Generate the simplified query By $S$
\STATE Compute confidence score $s_x$
\IF{$s_x>s_s$}
\STATE $P_S$.insert ({$q'_r,q_s,s_x$})
\ENDIF
\ENDFOR
\FOR{$q_r\in U_R$}
\STATE $q'_s\leftarrow$ Generate the rewritten query By $R$
\STATE Compute confidence score $s_y$
\IF{$s_y>s_r$}
\STATE $P_R$.insert ({$q_r,q'_s,s_y$})
\ENDIF
\ENDFOR
\STATE $U_S\leftarrow U_S\backslash P_S$, $U_R\leftarrow U_R\backslash P_R$, $P\leftarrow P_S\cup P_R$
\STATE $D_{Aug}\leftarrow Aug(D,P)$
\STATE Train $S$ and $R$ on $D_{Aug}$
\ENDWHILE
\end{algorithmic}
\end{algorithm}

\section{Limitations and Future Works}
\label{appendix:2}
Although our model has shown effectiveness in the CQR task, one drawback is that, the quality of the unlabeled data is vital to the model performance. How to choose hyperparameters such as the confidence threshold of the selectors is important but tricky. In addition, for the page limit, our work focus on the query rewrite mostly on the NLP level, while how much this paradigm will benifit  the conversational information retrieval system is still underexplored. Moreover, this Rewriter/Simplifier system can be adapted to other generative tasks, where in this work we only focus on the query write task. In our future work, we will work on exploring the co-training paradigm under other scenarios. We'll be exploring how co-training can be applied specifically for other conversational IR scenarios, ultimately enhancing user experiences and satisfaction.


\end{document}